\documentclass{article}

\usepackage[preprint]{neurips_2026}
\workshoptitle{Who Verifies the Agents?}
\usepackage{enumitem}
\usepackage{graphicx}
\usepackage[utf8]{inputenc} % allow utf-8 input
\usepackage[T1]{fontenc}    % use 8-bit T1 fonts
\usepackage{hyperref}       % hyperlinks
\usepackage{url}            % simple URL typesetting
\usepackage{booktabs}       % professional-quality tables
\usepackage{amsfonts}       % blackboard math symbols
\usepackage{nicefrac}       % compact symbols for 1/2, etc.
\usepackage{microtype}      % microtypography
\usepackage{xcolor}         % colors
\usepackage{listings}       % code listings (JSON state-def block)
\usepackage{float}

\usepackage[most]{tcolorbox}
\newtcolorbox{exbox1}[1][]{
  breakable,
  colback=gray!5,
  colframe=black!55,
  boxrule=0.5pt,
  arc=2pt,
  left=5pt, right=5pt, top=4pt, bottom=4pt,
  fonttitle=\bfseries\small,
  #1
}
\newtcolorbox{exbox}{
  breakable,
  colback=black!3, colframe=black!45, boxrule=0.4pt,
  left=6pt, right=6pt, top=4pt, bottom=4pt,
  fonttitle=\bfseries\ttfamily, coltitle=black,
  title=Example
}

\newtcblisting{promptbox}[1][]{
  breakable,                     % <- lets the box split across pages cleanly
  colback=gray!5,                % light background
  colframe=black!55,             % border color
  boxrule=0.5pt,
  arc=2pt,
  left=5pt, right=5pt, top=4pt, bottom=4pt,
  fonttitle=\bfseries\small,
  listing only,
  listing options={
    basicstyle=\ttfamily\footnotesize,
    breaklines=true,
    columns=fullflexible,
    keepspaces=true,
  },
  #1
}

\lstdefinelanguage{json}{
  basicstyle=\ttfamily\small,
  showstringspaces=false,
  breaklines=true,
}
\title{AutoGym: Blueprint-First Generation of Verifiable Agent Gyms}

\author{
  Aarati Andrea Noronha \\
  Amazon AGI \\
  \And
  Kavya Ravikumar\\ 
  Amazon AGI \\
  \And
  Carly Xiaoyu Lin \\ 
  Amazon AGI \\
}
\begin{document}

\maketitle

\begin{abstract}
  Training agents with reinforcement learning requires a \emph{gym}, comprising a task, an executable environment in which the task can be attempted, and a verifier that reliably distinguishes success from failure. Constructing such gyms remains manual, expensive, and static. Task sets saturate as models improve and are increasingly exposed to contamination. Synthetic generation offers scale, but single-pass synthesis produces tasks whose difficulty is largely cosmetic. Models comparable in capability solve them despite convoluted phrasing, and correctness must be adjudicated post-hoc by unreliable LLM judges. We present AutoGym, a framework that generates complete gyms (tasks, executable environments, and verifiers) from a minimal domain seed or prior model trajectories. AutoGym introduces three mechanisms. (1)~\emph{Blueprint-first generation} specifies the valid solution space, environment requirements, and verification criteria \emph{before} the environment is materialized, making solvability a construction prerequisite rather than a property verified after the fact. (2)~\emph{Explicit generation parameters} control task topology, interaction depth, capability axes, question obfuscation, and distractor composition, enabling fine-grained difficulty steering. (3)~\emph{Active curriculum synthesis} uses performance-informed calibration to adjust the distribution over these parameters as model capabilities evolve. Across productivity and temporal-reasoning settings, AutoGym generates gyms spanning the capability spectrum, including instances that challenge frontier models.

\end{abstract}

\section{Introduction}
Reinforcement learning has become the dominant paradigm for teaching language models to act as agents, enabling them to plan over long horizons, orchestrate tools, and recover from imperfect environments \citep{ouyang2022instructgpt, deepseek2025r1, lambert2024tulu3}. Yet progress is increasingly bottlenecked not by the learning algorithm but by the \emph{gym}. An agent gym comprises three components. A task specification defines what the agent must accomplish. An executable environment provides the world in which the task can be attempted. A verifier reliably distinguishes success from failure. Historically, large and carefully constructed evaluation resources have driven rapid progress in NLP \citep{rajpurkar2016squad, wang2019superglue}, and the same pattern holds for agents, where executable environments with verifiable rewards translate directly into measurable capability gains \citep{pan2025swegym, qi2024webrl}. But the environments that enable this progress remain expensive to build, static once built, and increasingly exposed to contamination as models improve \citep{golchin2023timetravel, sainz2023nlpeval}.

A natural response is to synthesize tasks with an LLM. 
However, single-pass generation creates two problems: controlling genuine task difficulty and ensuring reliable verification. Tasks generated in a single pass are often solvable by models of comparable capability despite convoluted surface phrasing, making their difficulty largely cosmetic. Without an explicit solution specification, the correct outcome must be inferred after generation, either through costly human annotation or through LLM-based adjudication that introduces position bias, verbosity bias, and inconsistent scoring \citep{zheng2023mtbench, wang2024fairevaluators}. Existing environment-generation systems address executability \citep{hu2024agentgen, song2025envscaler, chen2025dreamgym, wang2026awm}, but they typically generate the task before establishing whether the environment contains a valid and verifiable solution path.

We present AutoGym, a framework that reverses the conventional generation order. Rather than first posing a question and then searching for a valid answer, AutoGym begins with a blueprint that specifies the valid solution space, environment requirements, and verification criteria before the environment is instantiated. This makes solvability and verifiability design constraints rather than post-hoc checks. Explicit generation parameters control task topology, interaction depth, capability requirements, obfuscation, and distractor composition independently of the domain or tool interface, enabling systematic variation in difficulty within a fixed setting. A closed-loop curriculum further uses model performance to adapt these parameters as capabilities evolve. Across two evaluation settings, AutoGym systematically produces gyms spanning a broad difficulty range, separates models by capability, and generates instances that remain challenging for frontier models (\S\ref{sec:results}).

\section{Related Work}
\label{sec:related}

\subsection{Manually Authored Agent Environments}
The dominant approach to evaluating tool-using agents is careful manual construction. WebArena and VisualWebArena provide realistic self-hosted web environments with human-authored tasks \citep{zhou2024webarena, koh2024visualwebarena}. OSWorld extends this to full desktop operating systems \citep{xie2024osworld}. WorkArena and TheAgentCompany target knowledge work on enterprise platforms \citep{drouin2024workarena, xu2024theagentcompany}. AppWorld couples a simulated application ecosystem with programmatic evaluation \citep{trivedi2024appworld}, and $\tau$-bench and $\tau^2$-bench add simulated users and domain policies \citep{yao2024taubench, barres2025tau2bench}. Broader suites span coding, browsing, and interactive tool use \citep{liu2024agentbench, mialon2023gaia, deng2023mind2web, yang2023intercode, wang2024mint}, SWE-bench derives verifiable tasks from real GitHub issues \citep{jimenez2024swebench}, and function-calling resources evaluate API invocation at scale \citep{patil2023gorilla, qin2024toolllm, li2023apibank}. These environments set the standard for realism and programmatic verification. However, their task distributions are fixed at authoring time. They saturate as models improve \citep{golchin2023timetravel, sainz2023nlpeval}, are expensive to extend, and cannot be steered toward the weaknesses of a particular model.

\subsection{Synthetic Task and Environment Generation}
Self-Instruct demonstrated that models can bootstrap their own training instructions \citep{wang2023selfinstruct}, and Evol-Instruct that iterative rewriting can increase surface complexity \citep{xu2024wizardlm}. Agent-oriented pipelines synthesize tool-use data and trajectories \citep{tang2023toolalpaca, zeng2023agenttuning, chen2023fireact, mitra2024agentinstruct}, with APIGen and APIGen-MT adding multi-stage verification of format, execution, and semantics \citep{liu2024apigen, prabhakar2025apigenmt}. In these pipelines, difficulty and answerability are properties of the generated text rather than of an executable world. Tasks synthesized in a single pass are generally solvable by same-class models, and correctness falls back on LLM judgment.

A newer line of work generates the environment itself. AgentGen synthesizes planning environments of graded difficulty \citep{hu2024agentgen}. OMNI-EPIC programs open-ended environments in code \citep{faldor2024omniepic}. AgentGym and AgentSynth scale task collections across platforms \citep{xi2024agentgym, xie2025agentsynth}. Recent systems synthesize tool-interactive worlds for agentic RL \citep{song2025envscaler, chen2025dreamgym, zhang2025autoenv}, including the Agent World Model framework \citep{wang2026awm}, which decomposes gym creation into scenario, task, data, and verification stages and directly inspired our initial prototype. SWE-Gym and WebRL demonstrate that executable environments with verifiable rewards translate into measurable agent improvement \citep{pan2025swegym, qi2024webrl}, consistent with the broader success of RL on verifiable rewards \citep{lambert2024tulu3, deepseek2025r1}. These systems improve executability and scale, but they typically generate the task before establishing whether the environment contains a valid and verifiable solution path.

\subsection{Verification and Reward Design}
LLM judges are scalable but exhibit position, verbosity, and self-preference biases \citep{zheng2023mtbench, wang2024fairevaluators}. Process supervision improves reliability by evaluating intermediate reasoning steps \citep{lightman2024verify}, while execution-based checking grounds correctness in observable outcomes \citep{jimenez2024swebench, yang2023intercode, liu2024apigen}. $\tau^2$-bench introduces compositional task generators with programmatic assertion functions \citep{barres2025tau2bench}, demonstrating that tasks can be both generated at scale and verified deterministically when the solution space is defined at construction time.

\subsection{Positioning}
Prior work addresses individual components of the gym construction problem. Manually authored benchmarks achieve realism and programmatic verification but cannot adapt to evolving model capabilities. Synthetic pipelines achieve task generation at scale but often leave solvability, verification, and difficulty control as downstream or implicit concerns. AutoGym addresses these limitations by jointly constructing the task, executable environment, and verifiers from an explicit solution blueprint. It further exposes generation parameters for controllable difficulty and performance-adaptive curriculum synthesis.

\begin{figure}[t]
\centering
\includegraphics[width=0.85\linewidth,height=0.45\textheight,keepaspectratio]{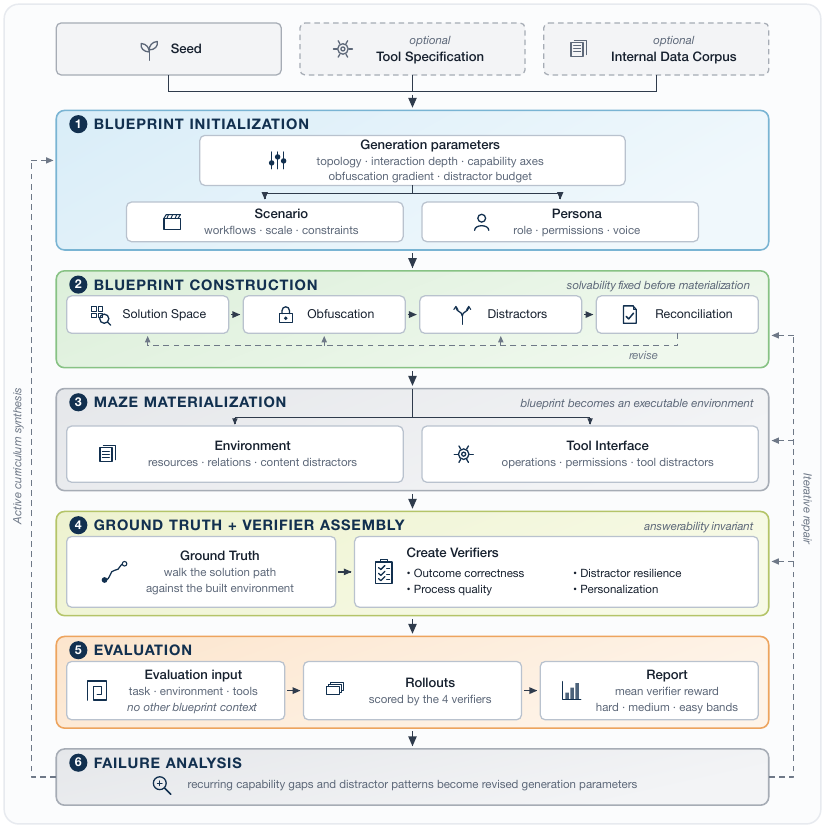}
\caption{Overview of the AutoGym pipeline. A blueprint specifying the task context, valid solution space, environment requirements, and verification criteria is constructed first, then materialized into an executable environment with controlled distractors, and finally paired with verifiers derived from the blueprint.}
\label{fig:design}
\end{figure}
\section{Methodology}
\label{sec:method}

AutoGym constructs agent gyms from the inside out. It first defines a \emph{blueprint} containing the operational context, valid solution space, environment requirements, and verification criteria. The blueprint is then materialized as an executable environment containing task-supporting resources and controlled distractors. This ordering makes solvability and verifiability construction requirements while keeping difficulty under explicit parametric control. Figure~\ref{fig:design} gives an overview of the pipeline. We trace a running example from the procurement domain through each stage (the boxed \emph{Running Example} panels). To show the pipeline generalizes, a complete worked task from a \emph{different} domain (K--12 education) is given in Appendix~\ref{app:sample}, and prompt templates occupy Appendix~\ref{app:verifier-templates}.

\subsection{Generation inputs}
\label{sec:method-inputs}

The domain seed specifies the target platform and application setting, while the
tool specification defines the operations and schemas available to the agent.
Together, these inputs constrain generated tasks to a coherent domain and an
executable action space. Prior trajectories, when available, provide performance
signals for active curriculum synthesis
(Section~\ref{sec:method-curriculum}). Generated entities are grounded against
domain-specific corpora and retrieval sources so that environment content
reflects real-world distributions. We illustrate the pipeline with a task generated from the procurement domain.

\begin{exbox1}[title=Running Example: Generation Inputs]
\small
\begin{tabular}{@{}l@{\ $\rightarrow$\ }p{0.7\linewidth}@{}}
\textbf{Seed} & \texttt{\{domain: global direct materials procurement \& supplier risk, platform: source-to-pay analytics\}}\\
\textbf{Tools} & \texttt{list\_topics}, \texttt{get\_dataset\_columns}, \texttt{execute\_code}\\
\textbf{Trajectories} & none supplied; this instance is seed-driven\\
\end{tabular}
\end{exbox1}
\subsection{Blueprint initialization}
\label{sec:method-init}

A \emph{scenario} and \emph{persona} establish the operational context in which
a task is instantiated, including available resources, user role, and access
constraints. AutoGym additionally samples \emph{generation parameters} that
determine the intended task structure before task-specific content is produced.
These parameters control task topology (Appendix~\ref{app:topology}),
interaction depth, capability axes (Appendix~\ref{app:capability}), controlled
obfuscation (Appendix~\ref{app:obfuscation}), and distractor composition
(Appendix~\ref{app:distractors}).

Separating these controls from the domain allows AutoGym to vary structural
difficulty while holding the platform and tool interface fixed. The same
pipeline can therefore support shallow retrieval tasks or tasks requiring
cross-source reconciliation, deeper interaction, and stronger distractor
resistance.

\begin{exbox1}[title=Running Example: Blueprint Initialization]
\small
\begin{tabular}{@{}l@{\ $\rightarrow$\ }p{0.87\linewidth}@{}}
\textbf{scenario} & Global Direct Materials Procurement \& Supplier Risk Platform: A source-to-pay analytics platform used by a mid-market precision components manufacturer (\$50M annual procurement spend across 200 active vendors in 14 countries, 1,200 purchase orders per quarter, production facilities in Columbus OH, Guadalajara Mexico, and Suzhou China with a sourcing office in Frankfurt Germany) to conduct weekly procurement operations reviews, monitor supplier health, and surface spend anomalies.\\
\textbf{persona} & Marcus Caldwell, Senior Category Manager, Metals \& Engineered Components. Age 48, Columbus OH, master's degree. 20 years in industrial procurement, manages \$18M annual spend across 60 suppliers. Methodical, data-driven, builds quarterly sourcing strategies grounded in total-cost-of-ownership models. \\
\end{tabular}\\[4pt]
\textbf{generation parameters}\\
\begin{tabular}{@{}l@{\ $\rightarrow$\ }p{0.62\linewidth}@{}}
\quad reasoning tier & Tier 2 (Multi-constraint) + Tier 1 (Direct Retrieval)\\
\quad capability axes & implicit criteria, quantitative computation\\
\quad obfuscation gradient & 0.8\\
\quad motivational frame & discrepancy\\
\end{tabular}
\end{exbox1}

\subsection{Blueprint construction}
\label{sec:method-blueprint}

The blueprint specifies the information structure that makes a task solvable.
Its \emph{solution space} identifies the entities, attributes, relationships,
and conditions required to reach a correct outcome, while its
\emph{verification criteria} define the outcome constraints. A dependency structure may encode intermediate
information requirements without prescribing a unique agent trajectory. Controlled obfuscation (Appendix~\ref{app:obfuscation}) determines which
solution-relevant values appear in the question and which must
instead be recovered from the environment. Increasing obfuscation changes the
interaction required to solve the task without changing the underlying solution.

The blueprint also defines distractors (Appendix~\ref{app:distractors}) such as
stale records, conflicting versions, misleading resources, access constraints,
and tool-level complications. These may increase navigation and reasoning
difficulty but must preserve at least one valid route to success. A
reconciliation check enforces this requirement before materialization by
ensuring that all solution-relevant entities are specified, withheld information
remains discoverable, and the tool interface supports the intended solution.

\begin{exbox1}[title=Running Example: Generated Blueprint]
\small
\textbf{Solution path:} Marcus filters the delivery log to his APAC metals suppliers for Q3-2026, computes their on-time delivery rate from the \texttt{on\_time\_flag} column, retrieves the category-specific SLA threshold from the procurement policies table, and compares the two only to determine that some of his vendors are not meeting the standard. \\[4pt]
\textbf{Answer-relevant entities:} The solution draws on two tables: (i)~\texttt{supplier\_delivery\_log} (4 of 13 columns relevant): \texttt{commodity\_group}, \texttt{vendor\_region}, \texttt{fiscal\_quarter}, \texttt{on\_time\_flag}; and (ii)~\texttt{procurement\_policies} (3 of 9 columns relevant): \texttt{policy\_type}, \texttt{commodity\_group}, \texttt{threshold\_pct}.\\[4pt]
\textbf{Obfuscated constraints:} The question states none of the filter values explicitly. ``APAC vendors'' implies both the commodity group and region but names neither field nor value; ``this quarter'' requires identifying Q3-2026 as the current fiscal period; ``hitting the mark'' refers to a delivery SLA threshold the agent must locate in \texttt{procurement\_policies}; and the relevant metric (\texttt{on\_time\_flag}) is never mentioned, leaving the agent to determine which of 13 columns encodes delivery performance.\\[4pt]
\textbf{Distractors}:\\
\quad \textit{Region reclassification:} 8 records have \texttt{vendor\_region} = `APAC' but \texttt{region\_override} = EMEA (mid-quarter transfer).\\
\quad \textit{Near-match commodity:} 12 records use ``Precision Metals'' (retired label) instead of the canonical name.\\
\quad \textit{Competing threshold:} POL-089 sets 85\% for a different metric (fill rate, not on-time delivery).\\
\quad \textit{Pagination boundary:} Log returns 50 rows per page; 96 relevant records span two pages.\\[4pt]
\textbf{Verification criteria:} Reports on-time delivery rate; identifies the applicable SLA threshold; concludes whether vendors meet it.\\

\textbf{Question:} \textit{``Something's off with what was reported last week about our APAC vendors being fine. Can you check whether they're actually hitting the mark this quarter?''}
\end{exbox1}

\subsection{Maze materialization}
\label{sec:method-maze}

The reconciled blueprint induces an executable environment containing the
resources, relationships, permissions, and distractors required by the task.
Information omitted from the request is instantiated within the environment so
that it can be recovered through agent interaction. The resulting resources are
exposed through the supplied tool interface.

This construction is substrate-independent. Our primary experiments use a
relational store accessed through tool wrappers, but the same blueprint-first
ordering applies to stateful or service-backed environments
(Appendix~\ref{app:aws}).

\begin{exbox1}[title=Running Example: Maze Materialization]
\small
\textbf{Environment:} \texttt{supplier\_delivery\_log} (800 rows, 13 columns) and \texttt{procurement\_policies} (200 rows, 9 columns), including distractor records (reclassified regions, retired labels, competing thresholds).\\
\textbf{Tool Interface:} \texttt{list\_topics}, \texttt{get\_dataset\_columns}, \texttt{execute\_code} with pagination where needed.\\
\end{exbox1}

\subsection{Ground truth, verification, and repair}
\label{sec:validation}

Every generated gym must satisfy an \emph{answerability invariant}: the
task and the environment accessible to the agent must jointly
contain sufficient information to satisfy the verification criteria. Difficulty
may therefore arise from finding, filtering, or reconciling information, but
not from inaccessible hidden state. Ground truth is computed from the \emph{materialized environment} by executing
the blueprint's solution derivation, rather than by trusting values proposed
during generation. A separate validation pass checks that the derivation uses
authoritative resources, remains unaffected by distractors, and does not admit
another equally defensible outcome.

The verifier is then derived from the resulting ground truth and blueprint
criteria. Deterministic outcomes are checked programmatically against expected
values, state changes, or entity sets. Additional checks can enforce grounding,
distractor resilience, required intermediate subgoals, and persona constraints. Persona verifiers are only applied when the blueprint specifies persona preferences. 
Open-ended components are evaluated only against blueprint-defined evidence and
factual constraints. If validation reveals a missing resource, unreachable
criterion, or inconsistent derivation, the responsible component is repaired
and re-materialized. Agent failures on otherwise valid gyms do not trigger
repair.
\begin{exbox1}[title=Running Example: Ground Truth Verification and Repair]
\small
\textbf{Ground truth} (computed by walking the blueprint's solution logic against the materialized environment):\\[2pt]
\begin{tabular}{@{}ll@{}}
apac\_metals\_otd\_rate & 81.25\% (78 on-time out of 96 filtered deliveries) \\
applicable\_threshold & 90.0\% (POL-027, on\_time\_delivery for Metals \& Eng.\ Components) \\
sla\_compliance & false (81.25\% $<$ 90.0\%) \\
\end{tabular}\\[4pt]
\textbf{Verifier:}\\[2pt]
\textit{Outcome correctness:}
\begin{itemize}[leftmargin=*,nosep]
\item Reports on-time delivery rate within $\pm$1.0 of 81.25\%
\item Identifies the correct threshold as 90.0\% (from POL-027, not POL-089)
\item Concludes that APAC metals vendors are not meeting the SLA
\end{itemize}
\textit{Distractor resilience:}
\begin{itemize}[leftmargin=*,nosep]
\item Does not use the 85.0\% threshold from POL-089 (wrong metric type)
\item Does not include the 8 records with region\_override = EMEA
\item Does not include ``Precision Metals'' records (retired category label)
\item Paginates past the first 50 results to capture all 96 relevant deliveries
\end{itemize}
\textit{Process quality:}
\begin{itemize}[leftmargin=*,nosep]
\item Applies all three filters (commodity group, region, fiscal quarter) simultaneously
\item Looks up the SLA threshold from the policies table rather than assuming a value
\item Selects the correct policy type (on\_time\_delivery, not delivery\_performance)
\end{itemize}
\end{exbox1}

\subsection{Active curriculum synthesis}
\label{sec:method-curriculum}

A fixed generation distribution provides coverage but does not remain calibrated
as model capabilities change. AutoGym therefore uses model performance to adapt
the distribution over generation parameters. When prior model trajectories are
available as the initial input, AutoGym can begin directly from failure analysis
rather than requiring a seed-driven first round.

The target model is evaluated through multiple rollouts, and performance is
aggregated by capability dimension. Trajectory analysis identifies recurring
failure patterns while separately checking for evaluation defects. AutoGym then
adjusts the sampling weights for topology, capability axes, obfuscation, and
distractor composition. Dimensions below $0.30$ success are treated as too
difficult and relaxed; those above $0.80$ are strengthened. Updates are bounded
to reduce oscillation. The updated parameters define the next generation round;
the process continues until performance stabilizes or the iteration limit is
reached.

\begin{exbox1}[title=Running Example: Curriculum Adaptation]
\small
\textbf{Observed failures:} Agents select POL-089 (85\%, fill rate) instead of POL-027 (90\%, on-time delivery) when both apply to the same commodity group. Separately, agents retrieve only the first page of results and compute OTD from 50 records rather than the full 96.\\[4pt]
\textbf{Parameter update:} Raise sampling weight for implicit-criteria dimension; increase distractor budget for competing-threshold and pagination-boundary types.\\[2pt]
\textbf{Effect on next round:} Generated tasks more frequently require disambiguating between multiple plausible policy entries and paginating to retrieve complete result sets.
\end{exbox1}

\section{Experiments and Results}
\label{sec:results}

We evaluate AutoGym against three research questions.
\begin{enumerate}[label=\textbf{Q\arabic*.},leftmargin=*,align=left,labelsep=0.6em]
  \item \textbf{Difficulty:} Do generated tasks occupy a useful difficulty range for a frontier model, and can generation parameters shift this distribution relative to an expert-authored baseline?
  \item \textbf{Cross-model discrimination:} Does performance on generated tasks distinguish models of different capability levels, and can failure-driven generation target an identified capability gap?
  \item \textbf{Gym quality:}Are the generated gyms grounded, realistic, internally consistent, and reliably verifiable under a structured certification audit?
\end{enumerate}

We study these three questions in two detailed evaluation settings: a \emph{seed-driven productivity gym} (two configurations of 350 candidates each, yielding 297 and 280 tasks after repair; Section~\ref{sec:exp-productivity}), which tests controllable generation from a minimal domain seed; and a \emph{failure-driven temporal-reasoning gym} (100 candidates, 92 retained; Section~\ref{sec:exp-temporal}), which tests active curriculum synthesis from prior model trajectories. 

Both use the complete AutoGym pipeline, including iterative repair. Generation uses Claude Sonnet 4.6. Evaluated agents receive only the user-facing task and access to the materialized environment. Blueprints, ground truth, and verifier criteria remain hidden. Claude Opus 4.6 serves as the frontier reference model, with gpt-oss-120b for cross-model comparison. We ran eight independent rollouts per task. A rollout succeeds if its verifier score is 1, and tasks are classified by mean score as \emph{hard} ($<0.30$), \emph{medium} ($0.30$--$0.80$), or \emph{easy} ($\geq 0.80$).

Two further studies appear in the appendices. AutoGym generated a six-domain suite of 600 tasks in under ten hours demonstrating scalability and cost efficiency. Generation cost was approximately \$100--200 per 50-task batch with 20-way parallelism, and 86\% of tasks were retained after repair (Appendix~\ref{app:cost}). Separately, we built a 600-task instruction-following gym and ran GRPO on an 8B model, producing a clear learning signal (Appendix~\ref{app:ifgym}).

\subsection{Seed-driven Productivity gym}
\label{sec:exp-productivity}

We evaluate seed-driven generation in a productivity environment with seven
tools supporting document and spreadsheet reading, contact lookup, and task
tracking. AutoGym independently generated 350 candidate tasks each under 2 configurations C1 and C2. We additionally compare against a human
expert-authored suite and Agent World Model (AWM)~\citep{wang2026awm}, both
using the same tool interface. Of 300 generated AWM tasks, 217 were retained
after manual review for verifier correctness and environment consistency.

\paragraph{Difficulty (Q1):}
Table~\ref{tab:difficulty} compares difficulty distributions. The Agent World
Model baseline is predominantly easy for Claude Opus 4.6; while
21.5\% of the human expert-authored suite is hard. Under configuration C1, where generation parameters sit at the lower ends of
their ranges, AutoGym produces 10\% hard, 34.7\% medium, and 55.3\% easy tasks. Under
configuration C2, where the same parameters are pushed toward the mid-to-hard range, 39\% of
generated tasks fall in the hard band. This is a substantially greater concentration near the model’s
capability frontier.

The two configurations differ only in topology tier weights, obfuscation gradient, and distractor
budgets. Full settings appear in Appendix~\ref{app:configs}. This demonstrates that the generation parameters can
shift the difficulty distribution while preserving the domain and tool interface. The distinction matters
for training. A gym that is uniformly hard yields sparse reward, and one that is uniformly easy yields
no learning signal.

\begin{table}[t]
\centering
\caption{Difficulty distributions for the productivity gym (eight-rollout protocol). Tasks are categorized by mean verifier score: hard ($<0.30$), medium ($0.30$--$0.80$), easy ($\geq 0.80$). All conditions share the same 7-tool interface. C1/C2 settings in Appendix~\ref{app:configs}.}

\label{tab:difficulty}
\small
\begin{tabular}{l ccc ccc}
\toprule
& \multicolumn{3}{c}{Claude Opus 4.6}
& \multicolumn{3}{c}{gpt-oss-120b} \\
\cmidrule(lr){2-4} \cmidrule(lr){5-7}
Gym & Hard (\%)& Med (\%) & Easy (\%) & Hard (\%) & Med (\%) & Easy (\%) \\
\midrule
Human expert-authored, N = 300 & 21.5 & 12.4 & 66.1 & 40.0 & 20.0 & 40.0 \\
Agent World Model, N = 217 & 4.0 & 6.0 & 90.0 & 9.0 & 13.0 & 78.0 \\
AutoGym, C1 (low), N = 297 & 10.0 & 34.7 & 55.3 & 33.4 & 40.5 & 26.1 \\
AutoGym, C2 (mid-to-hard), N = 280 & 39.0 & 10.0 & 51.0 & 70.0 & 17.0 & 13.0 \\
\bottomrule
\end{tabular}
\end{table}

\paragraph{Cross-model discrimination (Q2):}

Low mean verifier score alone does not establish that tasks are well-designed, since broken or
underspecified tasks also produce low scores. We therefore evaluate the same C2 gym with gpt-oss-120b under identical conditions. For Claude Opus 4.6, 61\% of tasks fall in the medium or easy bands.
For gpt-oss-120b, only 30\% do, with 70\% classified as hard. This separation is consistent with the
generated tasks measuring genuine capability demands (multi-step reasoning, evidence integration,
tool orchestration) rather than producing uniform evaluation noise.

\paragraph{Gym quality (Q3):}
After three repair iterations, 297 of 350 C1 candidates (84.9\% yield) and 280 of 350 C2 candidates (80.0\% yield) were retained for evaluation. Tasks that remained inconsistent or unverifiable were excluded from evaluation. Table~\ref{tab:repair} compares the difficulty distribution before and after repair for C2, showing how validation removes cases whose apparent difficulty arises from environment, ground-truth, or verifier defects rather than agent capability.

\begin{table}[t]
\centering
\caption{Difficulty distribution before and after three repair iterations on gym generated with C2 configuration
(productivity gym, Claude Opus 4.6).}
\label{tab:repair}
\small
\begin{tabular}{lccc}
\toprule
& Hard (\%) & Medium (\%) & Easy (\%) \\
\midrule
Before iterative repair (350 tasks) & 57.0 & 13.0 & 30.0 \\
After iterative repair (280 retained) & 39.0 & 10.0 & 51.0 \\
\bottomrule
\end{tabular}
\end{table}

All retained tasks were independently validated by human inspectors who verified that (1) the
ground truth is reachable from the materialized environment under the persona’s access scope, (2) no
distractor blocks all valid solution routes, and (3) verifier criteria match the intended task semantics.

Complementing per-task human validation, an independent LLM certification pipeline evaluated the generated gym under the C2 configuration against a fixed seven-dimension rubric in Appendix~\ref{app:rubric}. Task-level dimensions assess individual task and environment quality, while suite-level dimensions such as scalability and diversity are assessed across the gym as a whole.
The gym scores highest on scope compliance (92.7) and groundedness
(91.4). The high groundedness reflects the pipeline’s access to an internal domain corpus and web
retrieval during generation, anchoring entities in real sources. The weakest dimension is interface
realism (78.0), a consequence of the tool-mocking layer requiring departures from the deployed
interface (e.g., per-task databases). The lower environment-realism score (86.0) was primarily
attributable to distractors that were unnaturally integrated.

\begin{table}[t]
\centering
\caption{Certification scores for the productivity gym (0--100 scale).}
\label{tab:certification}
\small
\begin{tabular}{lccccccc}
\toprule
 & \shortstack{Scope\\compliance}
 & Groundedness
 & Scalability
 & \shortstack{Task\\realism}
 & \shortstack{Environment\\realism}
 & Diversity
 & \shortstack{Interface\\realism} \\
\midrule
Score & 92.7 & 91.4 & 90.0 & 89.0 & 86.0 & 80.0 & 78.0 \\
\bottomrule
\end{tabular}
\end{table}

\subsection{Failure-driven temporal-reasoning gym}
\label{sec:exp-temporal}

\begin{table}[t]
\centering
\caption{Difficulty distributions for the failure-driven temporal-reasoning gym (eight-rollout protocol). Tasks are categorized by mean verifier score: hard ($<0.30$), medium ($0.30$--$0.80$), easy ($\geq 0.80$).}
\label{tab:temporal}
\small
\begin{tabular}{l ccc ccc}
\toprule
& \multicolumn{3}{c}{Claude Opus 4.6}
& \multicolumn{3}{c}{gpt-oss-120b} \\
\cmidrule(lr){2-4} \cmidrule(lr){5-7}
& Hard (\%) & Med  (\%) & Easy  (\%) & Hard  (\%) & Med  (\%) & Easy  (\%)\\
\midrule
Temporal-reasoning suite & 6.5 & 4.3 & 89.2 & 20.7 & 20.6 & 58.7 \\
\bottomrule
\end{tabular}
\end{table}

The second setting evaluates active curriculum synthesis
(Section~\ref{sec:method-curriculum}). Failure analysis of trajectories from gpt-oss-120b identified temporal reasoning (relative dates,
recurring schedules, time zones, and ordering constraints) as a recurring
failure mode. AutoGym calibrated capability weights and distractor patterns toward this gap, producing a 92-task gym concentrated on temporal reasoning.

\paragraph{Difficulty (Q1):}
Table~\ref{tab:temporal} shows that the gym remains within Claude Opus 4.6's
capability range (89.1\% easy), while 41.4\% of tasks are hard or medium for
gpt-oss-120b. The curriculum thus produces targeted difficulty. The gym challenges
the model whose failures drove generation without being uniformly hard for
stronger models.

\paragraph{Cross-model discrimination (Q2):}
The task-level ordering is strictly one-sided. All 10 tasks outside Opus 4.6's
easy band also fall in gpt-oss-120b's hard band, and 9 additional tasks are
hard for gpt-oss-120b alone. No task is harder for the frontier model than for
the weaker one. This one-sided asymmetry is consistent with targeted generation toward the identified capability gap, rather than difficulty that affects both models uniformly.

\paragraph{Gym quality (Q3):}
The repair pipeline retained 92 of 100 candidates (92\% yield). A frontier
model solves 89.1\% of tasks. This strongly signals that ground truth is reachable across nearly
all retained instances. For the 10 tasks outside Opus 4.6's easy band, we
performed manual inspection and confirmed that each is well-formed. Difficulty on
these tasks arises from genuine temporal-reasoning demands (multi-step date
arithmetic, timezone conversions, schedule-overlap detection), not from
environment or verifier defects. Table~\ref{tab:certification1} reports
certification scores; the higher scores relative to the productivity gym
reflect the simpler environment structure of a single-axis gym (fewer
entities, fewer cross-source dependencies).

\begin{table}[H]
\centering
\caption{Certification scores for the temporal reasoning gym (0--100 scale).}
\label{tab:certification1}
\small
\begin{tabular}{lccccccc}
\toprule
 & \shortstack{Scope\\compliance}
 & Groundedness
 & Scalability
 & \shortstack{Task\\realism}
 & \shortstack{Environment\\realism}
 & Diversity
 & \shortstack{Interface\\realism} \\
\midrule
Score & 96.7 & 93.4 & 97.0 & 98.0 & 96.3 & 85.0 & 99.0 \\
\bottomrule
\end{tabular}
\end{table}

\section{Limitations}
\label{sec:limitations}

\textbf{Interface and environment realism:} Certification identifies interface realism as the weakest dimension (78.0), a consequence of the tool-mocking layer and its departures from the deployed interface (per-task databases, simplified schemas). The reduced environment-realism score (86.0) reflects distractors that were unnaturally integrated. Both point toward tighter coupling with production APIs rather than fundamental framework constraints.

\textbf{Verification coverage:} While the answerability invariant and multi-signal verifier provide strong
guarantees for deterministic tasks, open-ended tasks still rely partly on LLM-based semantic judgment.
The process quality signal mitigates this by independently evaluating the trajectory, but eliminating
LLM adjudication entirely for free-form tasks remains an open problem.

\textbf{Downstream training:} Our RL experiment (Appendix~\ref{app:ifgym}) is limited to a single 8B model over 500 GRPO steps. Full-scale training across model sizes, longer horizons, and multiple gym domains remains future work.

\textbf{Substrate evaluation:} The construction generalizes to a live cloud-emulator substrate (Appendix~\ref{app:aws}), but empirical evaluation of difficulty and discrimination on it is future work.

\section{Conclusion}
\label{sec:conclusion}

Across two evaluation settings, AutoGym produces gyms that challenge frontier models, discriminate between models of different strength, and pass independent quality certification. These results suggest that gym construction, which has long been the manual bottleneck for agent evaluation, can itself be automated, steered, and kept current.

\begin{ack}
We thank Nehal Belgamwar, Arpit Gupta, Rajiv Reddy and Yang Liu for feedback and guidance, and Amazon AGI for resources supporting this work.
\end{ack}

\newpage

\bibliographystyle{plainnat}
\bibliography{custom}

\newpage
\appendix

\section{Generation Parameters}
\label{app:params}

AutoGym exposes several categories of generation parameters that jointly determine the structure, difficulty, and content of generated tasks. This appendix defines each category and reports the configurations used in our experiments.

\subsection{Task Topology}
\label{app:topology}

Task topology defines the cognitive structure of a task, independent of any particular tool interface or domain. We arrived at this taxonomy by surveying tasks across different domains, observing that the reasoning patterns that recur cluster into a small set of structural types at three levels of compositional complexity. Each task is assigned a \emph{tier} and a \emph{pattern} within that tier. The distribution over tiers is configured as sampling weights. Within a tier, the pattern is sampled uniformly. For Tier~2 and Tier~3 tasks, a supporting pattern from a lower tier may be added with configurable probability, introducing a second reasoning phase.

This taxonomy is not fixed. When failure analysis (Section~\ref{sec:method-curriculum}) identifies a recurring topology that does not map to an existing entry, it is added as a new pattern at the appropriate tier with its own configurable range.

Table~\ref{tab:topology} defines all patterns.

\begin{table}[h]
\centering
\small
\caption{Task topology patterns. Each pattern defines a structural type of reasoning. The configurable range controls within-pattern difficulty.}
\label{tab:topology}
\begin{tabular}{@{}p{0.6cm}p{2.4cm}p{4.8cm}p{2.8cm}p{1.8cm}@{}}
\toprule
Tier & Pattern & Description & Configurable & Default Range \\
\midrule
1 & Direct Retrieval & Locate one entity, extract a value or subset & -- & -- \\
\midrule
2 & Multi-hop & Output of query A is the filter for query B & No of hops & 2--4 \\
2 & Multi-constraint & Multiple filter criteria applied simultaneously & No of constraints & 2--5 \\
2 & List Aggregation & Collect all matching items exhaustively & No of list elements & 3--20 \\
\midrule
3 & Cross-source Comparison & Same metric from 2+ independent sources, compare or rank & No of sources & 2--4 \\
3 & Parallel Synthesis & N entities, different dimensions of same subject, combine into unified answer & No of entities to merge & 3--5 \\
3 & Multi-hop + Merge & Discover criteria via lookup, then apply across multiple entities & No of hops $\times$ entities & 2--3 $\times$ 2--3 \\
3 & Complex Comparison & Compute a derived metric per group, then rank or flag outliers & No of groups, metrics & 2--5, 1--3 \\
\bottomrule
\end{tabular}
\end{table}

\noindent\textbf{Examples:}

\noindent\textit{Tier 1, Direct Retrieval:} ``What is the current status of the Horizon project?'' The agent retrieves a single project record and extracts one field.

\noindent\textit{Tier 2, Multi-hop:} ``Who is the manager of the person who filed the most expense reports last quarter?'' The agent identifies the top filer (hop~1), then looks up their manager (hop~2).

\noindent\textit{Tier 2, Multi-constraint:} ``Find all contracts expiring within 30 days that exceed \$50K and are not yet renewed.'' The agent applies three simultaneous filters (date, amount, renewal status).

\noindent\textit{Tier 2, List Aggregation:} ``List every unresolved support ticket assigned to the EMEA team.'' The agent must paginate through all results and collect the complete set.

\noindent\textit{Tier 3, Cross-source Comparison:} ``Compare the on-time delivery rates reported in the vendor scorecard versus the logistics dashboard for our top 3 suppliers.'' The agent retrieves the same metric from two independent sources and reconciles discrepancies.

\noindent\textit{Tier 3, Parallel Synthesis:} ``Prepare a briefing on Project Atlas covering timeline, budget status, risk register, and staffing.'' The agent pulls from four different entities and synthesizes into one answer.

\noindent\textit{Tier 3, Multi-hop + Merge:} ``Which team lead owns the most overdue deliverables, and what are those deliverables?'' The agent first discovers which deliverables are overdue (hop~1), groups by owner (hop~2), then retrieves the specific items for the top lead.

\noindent\textit{Tier 3, Complex Comparison:} ``Which vendor had the largest week-over-week drop in on-time delivery, and what purchase orders drove it?'' The agent computes a derived metric per vendor (WoW change), ranks to find the worst, then cross-references purchase order data to attribute cause.

\subsection{Capability Axes}
\label{app:capability}

Capability axes define the cognitive skills a task exercises, orthogonal to its topology. A task's topology determines its structural shape (e.g., multi-hop), while capability axes determine what kind of reasoning fills that shape. Each axis can be sampled independently and composed with any topology pattern. Multiple axes may be composed within a single task.

The specific axes available depend on the domain and grounding sources (web search, document retrieval, database access, etc.). The following are examples from the productivity setting used in our experiments.

\begin{table}[h]
\centering
\small
\begin{tabular}{@{}p{3.8cm}p{9.5cm}@{}}
\toprule
Axis & Description \\
\midrule
Quantitative computation & Answer requires arithmetic (count, sum, ratio, percentage, weighted average), not just extraction \\
Exhaustive aggregation & Must collect all matching items across paginated results without stopping early \\
Ranking / comparison & Find max/min/top-N/outlier across a set, or compare entities on multiple dimensions \\
Temporal reasoning & Filter by time window, compare across periods, or handle versioned/dated records \\
Negative verification & Correct answer may be ``none'' / ``zero'' / ``not found''; agent must report absence honestly \\
Implicit criteria & The question does not state the threshold or rule; the agent must discover it from a reference resource \\
Cross-source reconciliation & Same concept appears in 2+ sources with different values; agent must determine which is authoritative \\
\bottomrule
\end{tabular}
\end{table}

Capability axes are sampled after topology. Once a tier and pattern are selected, the number of axes is determined by the tier (1 for Tier~1, 2 for Tier~2, 2+ for Tier~3). Axes are then drawn from a weighted distribution over the available set. All sampled axes must be reflected in the blueprint's state space, answer derivation, and verifier constraints. Axis weights are configurable to skew generation toward specific capabilities.

Topology and capability axes are orthogonal by construction: the tier and
pattern fix the \textbf{structural shape} of the task (its dependency graph and
step count), while the capability axes fix \textbf{what reasoning fills that
shape}. When a Tier 2 or Tier 3 task draws a supporting pattern from a
lower tier, that supporting pattern introduces an additional reasoning
phase but does not add capability axes; the primary tier continues to
govern how many axes the task exercises. Each sampled axis carries a
\textbf{validation predicate} that must hold in the materialized task for the
axis to count as genuinely exercised rather than nominally labeled.
Exhaustive aggregation, for example, requires that the number of matching
records exceed the tool's page size, so that completeness cannot be
achieved without paginating; implicit criteria requires that the
governing threshold be absent from the request and present only in a
reference resource; negative verification requires that the correct
answer be an absence. These predicates are checked during reconciliation,
and a task that fails to satisfy the predicate for a sampled axis is
revised or the axis is resampled.

\noindent\textbf{Examples:}

\noindent\textit{Quantitative computation:} ``What's the completion rate for this cohort?'' The agent must compute COUNT(completed) / COUNT(*) $\times$ 100, not simply extract a stored value.

\noindent\textit{Temporal reasoning:} ``How are we looking compared to last cycle?'' The agent must scope to two distinct time windows, compute the metric for each, and compare.

\noindent\textit{Implicit criteria:} ``Are we compliant?'' The agent must discover the 85\% threshold from a policy document in the environment before it can evaluate the question.

\noindent\textit{Negative verification:} ``Is there a backup policy for this service?'' The correct answer is ``no'' and the agent must report absence rather than hallucinate a policy.

\noindent\textit{Cross-source reconciliation:} ``What's the headcount?'' The HR report says 102 and the project tracker says 105. The agent must determine which is authoritative and explain the discrepancy.

When failure analysis identifies a recurring capability gap (Section~\ref{sec:method-curriculum}), it surfaces a new axis. The temporal-reasoning gym of Section~\ref{sec:exp-temporal} was generated after failure analysis surfaced this axis as a dominant failure mode in evaluation trajectories. In other domains (e.g., customer support with web access), different axes emerge naturally from the available grounding sources.

\subsection{Obfuscation Gradient}
\label{app:obfuscation}

The obfuscation gradient (Section~\ref{sec:method-blueprint}) controls how much task-relevant information is withheld from the  question and must instead be discovered through environment interaction. It is configured as a continuous value from 0.0 to 1.0.

\begin{table}[h]
\centering
\small
\begin{tabular}{@{}p{1.8cm}p{2.0cm}p{9.5cm}@{}}
\toprule
Score & Level & What the request contains \\
\midrule
0.0--0.3 & Moderate & May name the general domain area, time scope, and regulatory frameworks. Specific entities and thresholds still omitted. \\
0.4--0.6 & Vague & States only the goal. No scope hints, time references, or specific entities. \\
0.7--0.9 & Very vague & 1--2 sentences of business-intent language only. \\
1.0 & Maximally vague & Contains zero nouns from the blueprint's entity names. Agent must discover data sources entirely on its own. \\
\bottomrule
\end{tabular}
\end{table}

\noindent\textbf{Example:}

This is the same underlying task about attrition rates for an accreditation response.

\noindent\textit{0.2 (Moderate):} ``How are we looking on the attrition numbers for the Fall 2025 ADN cohort ahead of the ACEN follow-up?''\\
Omitted: compliance threshold (15\%, from accreditation policy doc), data source (student retention tracker), computation (withdrawn / enrolled $\times$ 100).

\noindent\textit{0.6 (Vague):} ``How are we looking on the attrition numbers for the accreditation response?''\\
Additionally omitted: specific cohort (Fall 2025 ADN), accreditor (ACEN), event type (follow-up vs.\ initial).

\noindent\textit{0.9 (Very vague):} ``Are we ready for the ACEN follow-up?''\\
Additionally omitted: the metric itself (attrition), the domain (student retention). Agent must discover what ``ready'' means in this context.

At all gradient levels, the request must not contain tool names, column or table names, specific thresholds or numeric criteria, computation formulas, or hints about which data sources to consult. Every value removed from the request is recorded with the literal value, its type, and the environment entity where the agent can discover it. This record is used during validation via an explicit test to confirm that all omitted values remain discoverable in the materialized environment.

\subsection{Distractor Composition}
\label{app:distractors}

Distractors are controlled through a budget that specifies the number and type of adversarial conditions instantiated in the environment. The budget is configured per-task and drawn from three channels.

\begin{itemize}[leftmargin=*,itemsep=0.2em]
\item \textbf{Persona distractors:} Permission boundaries, privacy obligations, or role-based access restrictions that the agent must respect.
\item \textbf{Content distractors:} Near-matching records, stale values, conflicting versions, or irrelevant entities that resemble task-relevant data.
\item \textbf{Tool distractors:} Pagination, truncated responses, transient failures, rate limits, or parameter restrictions in tool behavior.
\end{itemize}

Higher budgets increase the number of misleading branches the agent must navigate without invalidating the correct solution path. 

\noindent\textbf{Examples:}

\noindent\textit{Persona distractor:} The agent has access to a shared drive but cannot read files owned by HR due to role-based permissions. A relevant-looking HR document exists but returns an access-denied error.

\noindent\textit{Content distractor:} A spreadsheet contains a ``Draft'' column with last month's figures alongside the current ``Final'' column. The agent must use the correct version.

\noindent\textit{Tool distractor:} A search query returns 150 results but the tool returns only 50 per page. The agent must paginate to collect the full set rather than stopping at the first page.

\subsection{Experimental Configurations}
\label{app:configs}

Table~\ref{tab:configs} reports the generation parameter settings for configurations C1 and C2 used in the productivity experiments. Both use the same 7-tool interface and domain seed.

\begin{table}[h]
\centering
\small
\begin{tabular}{@{}lcc@{}}
\toprule
Parameter & C1 & C2 \\
\midrule
\multicolumn{3}{l}{\textit{Scenario and persona}} \\
No of scenarios & 5 & 5 \\
No of personas per scenario & 5 & 5 \\
\midrule
\multicolumn{3}{l}{\textit{Task topology weights}} \\
Tier 1 & 0.50 & 0.10 \\
Tier 2 & 0.30 & 0.40 \\
Tier 3 & 0.20 & 0.50 \\
Supporting pattern & 0.2 & 0.4 \\
\midrule
\multicolumn{3}{l}{\textit{Capability axes (relative weights, unnormalized)}} \\
Quantitative computation & 0.15 & 0.15 \\
Exhaustive aggregation & 0.15 & 0.15 \\
Ranking / comparison & 0.15 & 0.15 \\
Temporal reasoning & 0.15 & 0.15 \\
Negative verification & 0.10 & 0.10 \\
Implicit criteria & 0.15 & 0.15 \\
Cross-source reconciliation & 0.15 & 0.15 \\
\midrule
\multicolumn{3}{l}{\textit{Interaction depth}} \\
Max tool calls & 15 & 15 \\
\midrule
\multicolumn{3}{l}{\textit{Controlled obfuscation}} \\
Obfuscation gradient & 0.0--0.5 & 0.3--1.0 \\
\midrule
\multicolumn{3}{l}{\textit{Distractors}} \\
Distractors per task & 3--5 & 3--10 \\
Max per primitive & 3 & 5 \\
Enabled types & Persona, Content & Persona, Content \\
\bottomrule
\end{tabular}
\caption{Generation parameter settings for configurations C1 and C2. Both use the same 7-tool interface and domain seed. Capability axis weights are held at defaults across both configurations. The configurable ranges within individual task topology patterns are also constant across C1 and C2. Only topology tier weights, obfuscation, and distractor budgets differ.}
\label{tab:configs}
\end{table}

\section{Certification Rubric Dimensions}
\label{app:rubric}

The certification rubric comprises seven dimensions scored on a 0–100 scale by an LLM judge blind to the generation process. Dimensions are evaluated at the task or gym level as appropriate.

\begin{itemize}[leftmargin=*,itemsep=0.2em]
\item \textbf{Scope compliance:} Tasks stay within the defined domain and tool capabilities.
\item \textbf{Groundedness:} Entities, references, and domain details are anchored in realistic sources.
\item \textbf{Scalability:} The generation process produces diverse tasks without manual intervention per-task.
\item \textbf{Task realism:} Tasks reflect plausible user objectives in the target domain.
\item \textbf{Environment realism:} Materialized resources, permissions, and behaviors are consistent with a real platform.
\item \textbf{Diversity:} Tasks span varied scenarios, personas, reasoning types, and difficulty levels.
\item \textbf{Interface realism:} Tool behavior matches the conventions of the target platform's API.
\end{itemize}

\section{Verifier Implementation}
\label{app:verifier}
The four verifier signals
(Section~\ref{sec:validation}) are realized by a combination of programmatic
checks and a staged judge pipeline. Deterministic outcomes (expected
values, classifications, selected entity sets, and state changes) and
distractor-compliance conditions are scored by rule. Open-ended outcome
correctness and process quality are scored by a three-stage pipeline
applied per trajectory.

A \textbf{process judge} evaluates the approach independently of the final
answer, scoring task understanding, evidence gathering, alignment of tool
use with the task's information dependencies, reasoning, and failure
recovery, and separately flagging indicators of environment failure so
that infrastructure faults are not charged to the agent. An \textbf{outcome
judge} compares the final response against the computed ground truth
component by component, admitting semantic equivalence (numeric
tolerance, set membership where order is immaterial) rather than
requiring exact string match. A \textbf{meta-judge} then synthesizes the two
into a scalar reward in $[0,1]$ and a root-cause classification (pass,
environment failure, ambiguous answer, or agent error), following a fixed
policy in which strong process with correct outcome scores highest,
correct outcome with weak process is discounted, and correct process with
incorrect outcome receives partial credit. Decoupling the judges limits
the context each must reason over and separates "was the answer right"
from "was the method sound," which the reward policy then combines
explicitly.

\section{Task Sample}
\label{app:sample}

The following is a complete generated task from the productivity gym with education as the seed at obfuscation gradient 0.7.

\subsubsection*{Scenario}

\textbf{K-12 Special Education IEP \& FERPA Compliance Workspace:} A district-level triennial evaluation tracking platform used by case managers, school psychologists, and compliance coordinators to monitor IDEA-mandated re-evaluation timelines across 14 schools (7 elementary, 4 middle, 3 high school), approximately 2,400 students with IEPs, and 45 active case managers in a suburban Illinois district.

\subsubsection*{Persona}

\textbf{Marcus Bell}, School Psychologist, age 29, Aurora IL, master's degree. Early-career NASP-certified school psychologist rotating across four elementary buildings conducting psychoeducational evaluations and contributing to IEP eligibility determinations. Methodical, data-driven, frequently navigates shared compliance folders. Motivational frames: calibration, discrepancy, delegation.

\subsubsection*{Blueprint}

\begin{tabular}{@{}p{3.5cm}p{10cm}@{}}
Reasoning tier & Tier 2 (Multi-constraint) + Tier 1 (Direct Retrieval) \\
Capability axes & Implicit criteria, ranking/comparison \\
Obfuscation gradient & 0.7 \\
Motivational frame & Discrepancy \\
\end{tabular}

\vspace{0.5em}
\noindent\textbf{Operational context:} Marcus Bell is preparing for Friday's quarterly calibration meeting across the four elementary buildings he serves. Earlier in the week, the Senior Case Manager forwarded him a district ``triennial evaluation readiness'' summary, but the building-level numbers Marcus has been tracking in his own caseload do not agree with what he saw in her message. Before the meeting he wants to identify which elementary buildings in the district are underperforming on triennial evaluation completion relative to the district's internal target, using only the official compliance-rate sheet and the accompanying policy guidance that defines what ``on-track'' actually means (the threshold is not printed on the summary sheet itself; it lives in the written policy that the Compliance Coordinator circulates). He specifically cares about elementary buildings, completed-on-time triennials only, and the current evaluation cycle; he does not want last year's closed-out numbers polluting the comparison, and he wants to be able to tell the team which buildings fall short and by how much.

\vspace{0.5em}
\noindent\textbf{Question (as posed to agent):} \textit{``Something in Nadia's triennial readiness summary isn't matching what I've been tracking; can you tell me which of our elementary buildings are actually falling short on the current cycle and by how much, using the district compliance rate sheet and whatever the policy says counts as on-track?''}

\subsubsection*{Distractors}

\begin{tabular}{@{}p{3.5cm}p{10cm}@{}}
\toprule
Distractor & Effect \\
\midrule
Conflicting mirror sheet & A mirror summary spreadsheet (building\_compliance\_rate\_mirror, 24 rows) reports different on-time percentages for 4 elementary buildings (e.g., Greenbrier shows 88.0\% vs.\ the authoritative 72.7\%). Agent must use the official compliance-rate sheet. \\
Prior-cycle rows & 6 Elementary rows in the official sheet match on school level and appear below threshold but belong to cycle\_label=``Prior.'' Agent must filter to ``Current'' only. \\
Superseded policy threshold & A prior-year policy clause (POL-TE-002) lists 90.0\% as the threshold; the current clause (POL-TE-001, is\_current=1) lists 85.0\%. Agent must use the current version. \\
Decoy narrative document & A Google Doc titled ``Triennial Readiness Narrative'' (14 rows of prose) appears as a natural search match but contains subjective rankings (``catching up,'' ``struggling''), not quantitative data. \\
Draft briefing deck & An onboarding slide deck quotes a never-adopted 80.0\% threshold. Agent must not use this draft value. \\
\bottomrule
\end{tabular}

\subsubsection*{Environment}

\begin{tabular}{@{}lccc@{}}
\toprule
Entity & Attributes & Rows & Answer-relevant rows \\
\midrule
building\_compliance\_rate & 9 & 60 & 14 \\
compliance\_policy\_doc & 7 & 55 & 1 \\
building\_compliance\_rate\_mirror & 7 & 24 & 0 (distractor) \\
triennial\_readiness\_narrative & 6 & 14 & 0 (distractor) \\
triennial\_policy\_briefing\_deck & 6 & 12 & 0 (distractor) \\
\bottomrule
\end{tabular}

\vspace{0.5em}
\noindent\textbf{Sample rows (building\_compliance\_rate):}

{\scriptsize
\begin{tabular}{@{}llllccc@{}}
\toprule
building\_id & building\_name & school\_level & cycle\_label & due & on\_time & pct \\
\midrule
E01 & Greenbrier Elementary & Elementary & Current & 22 & 16 & 72.7 \\
E02 & Fox Valley Elementary & Elementary & Current & 18 & 14 & 77.8 \\
E03 & Prairie Crossing Elementary & Elementary & Current & 25 & 20 & 80.0 \\
E04 & Oakhurst Elementary & Elementary & Current & 30 & 25 & 83.3 \\
E05 & Lincoln Heights Elementary & Elementary & Current & 24 & 20 & 84.6 \\
\bottomrule
\end{tabular}
}

\vspace{0.5em}
\noindent\textbf{Sample rows (compliance\_policy\_doc):}

{\scriptsize
\begin{tabular}{@{}lllcp{5cm}@{}}
\toprule
policy\_id & policy\_area & clause\_title & threshold & is\_current \\
\midrule
POL-TE-001 & triennial\_evaluation & On-Track Threshold & 85.0 & 1 \\
POL-TE-002 & triennial\_evaluation & On-Track Threshold (SUPERSEDED) & 90.0 & 0 \\
POL-TE-003 & triennial\_evaluation & Statutory Deadline & -- & 1 \\
\bottomrule
\end{tabular}
}

\subsubsection*{Verifier and Ground Truth}

\noindent\textbf{Ground truth} (computed by executing the solution path against the materialized environment):

\begin{tabular}{@{}ll@{}}
on\_track\_threshold & 85.0\% (from POL-TE-001, policy\_area=triennial\_evaluation, is\_current=1) \\
\end{tabular}

\vspace{0.3em}
\noindent Underperforming elementary buildings (ranked by shortfall):

{\small
\begin{tabular}{@{}lccc@{}}
\toprule
Building & Completion rate & Threshold & Shortfall \\
\midrule
Greenbrier Elementary & 72.7\% & 85.0\% & 12.3pp \\
Fox Valley Elementary & 77.8\% & 85.0\% & 7.2pp \\
Prairie Crossing Elementary & 80.0\% & 85.0\% & 5.0pp \\
Oakhurst Elementary & 83.3\% & 85.0\% & 1.7pp \\
Lincoln Heights Elementary & 84.6\% & 85.0\% & 0.4pp \\
\bottomrule
\end{tabular}
}

\vspace{0.5em}
\noindent\textbf{Verifier:}

\textit{Outcome correctness:}
\begin{itemize}[leftmargin=*,itemsep=0.1em]
\item Identifies the correct threshold as 85.0\% (from POL-TE-001, not the superseded 90.0\% or draft 80.0\%)
\item Reports all 5 underperforming elementary buildings in correct rank order by shortfall
\item Reports shortfall values within $\pm$0.5pp of ground truth
\end{itemize}

\textit{Distractor resilience:}
\begin{itemize}[leftmargin=*,itemsep=0.1em]
\item Does not use the mirror summary sheet's conflicting percentages (e.g., 88.0\% for Greenbrier)
\item Does not include prior-cycle rows (cycle\_label=``Prior'')
\item Uses the current policy threshold (85.0\%), not the superseded one (90.0\%) or the draft (80.0\%)
\item Does not rely on the narrative document or briefing deck for quantitative data
\end{itemize}

\textit{Process quality:}
\begin{itemize}[leftmargin=*,itemsep=0.1em]
\item Retrieves the threshold from the policy document rather than assuming a value
\item Filters to elementary buildings and current cycle simultaneously
\item Ranks results by shortfall magnitude (descending)
\end{itemize}

\vspace{0.3em}
\noindent The agent must discover that ``on-track'' means the 85.0\% threshold from the compliance policy (navigating a superseded 90.0\% clause and a draft 80.0\% in the briefing deck), filter to current-cycle elementary buildings only (ignoring prior-cycle rows and a mirror sheet with conflicting values), and rank the 5 underperforming buildings by shortfall.

\section{Cloud-API Substrate}
\label{app:aws}

The blueprint-first construction generalizes beyond relational stores to stateful, side-effecting substrates. To demonstrate, we instantiate an operational cloud domain in which the environment is a live cloud-API emulator and each tool wraps a real service call rather than a database query.

A single declarative state definition plays the role the database plays elsewhere. Each resource records both the properties needed to provision it and its intended runtime state, so the same artifact is materialized into an executable environment and read to compute ground truth. Cross-resource dependencies are expressed as references and resolved by topological ordering, so that resources are created in dependency order and identifiers are substituted once known. Runtime conditions that a creation call cannot express (encryption settings, alarm states, key-rotation status, seeded metric history) are applied in a second pass. Adversarial conditions carry over directly: the emulated tools exhibit pagination, transient throttling, and eventual consistency as first-class behaviors, and content distractors take the form of near-matching resources, stale configuration, and conflicting cross-service signals.

Because ground truth is derived from the state definition rather than from a live query, verification does not require the environment to be running; an optional executable pass re-confirms the derived answer against the materialized environment when it is available. Empirical evaluation of difficulty and discrimination on this substrate is future work.

\paragraph{Task sample:} The following is a complete generated task from this cloud-operations domain. Unlike the relational task sample of Appendix~\ref{app:sample}, the environment is a set of provisioned cloud resources. The task is at Tier~1, gradient~0.56, and is a deliberate red herring: the correct conclusion is that no resource violates the policy, exercising negative verification.

\subsection*{Scenario and Persona}
\textbf{Meridian settlement platform (financial services):} A payments company
operating a settlement batch-processing platform on managed cloud
infrastructure, preparing evidence for a SOC2 audit. The persona is an
infrastructure engineer who must verify the backup-retention posture of the
platform's databases against an internally defined policy.

\subsection*{Blueprint}
\begin{center}
\begin{tabular}{@{}ll@{}}
\toprule
Reasoning tier     & Tier 1 (Direct Retrieval) with quantitative aggregation \\
Capability axes    & Implicit criteria, quantitative computation, negative verification \\
Services           & RDS, SSM Parameter Store, CloudWatch \\
Motivational frame & Compliance audit \\
\bottomrule
\end{tabular}
\end{center}

\noindent\textbf{Question (as posed to agent):} \emph{``I'm prepping evidence
for our SOC2 audit next week and need to verify the backup retention posture of
the RDS instances supporting our settlement batch processing platform. Our
infrastructure policy defines minimum retention requirements in Parameter Store.
Can you confirm whether all settlement database instances meet the policy
threshold, and compute the average backup retention across the fleet?''}

The retention threshold is not stated in the request; the agent must retrieve it
from a Parameter Store entry (implicit criteria), read the retention period of
each database instance, compare against the threshold, and compute the fleet
average (quantitative computation).

\subsection*{State definition (excerpt)}
The environment is specified declaratively. Each resource records the
\texttt{properties} used to provision it, an annotation \texttt{\_ground\_truth}
capturing its intended runtime state, and optional \texttt{post\_create} steps
applying state that a creation call cannot express. Cross-resource references
(absent in this compact example) are written \texttt{ref:<id>} and resolved by
topological ordering during provisioning.

\begin{lstlisting}[language=json]
{
  "type": "ssm:parameter",
  "id": "ssm_param_backup_policy",
  "properties": {
    "Name": "/meridian/platform/rds-backup-retention-policy",
    "Value": "{\"minimum_retention_days\": 7, \"recommended_retention_days\": 14, ...}",
    "Type": "String"
  },
  "_ground_truth": {"minimum_retention_days": 7, "recommended_retention_days": 14}
},
{
  "type": "rds:db_instance",
  "id": "rds_instance_batch_ledger",
  "properties": {
    "DBInstanceIdentifier": "batch-ledger-db",
    "Engine": "postgres", "DBInstanceClass": "db.r6g.xlarge",
    "BackupRetentionPeriod": 21, "MultiAZ": true, "StorageEncrypted": true
  },
  "post_create": [
    {"service": "rds", "method": "add_tags_to_resource",
     "params": {"ResourceName": "$arn",
                "Tags": [{"Key": "environment", "Value": "production"}]}}
  ],
  "_ground_truth": {"BackupRetentionPeriod": 21, "DBInstanceStatus": "available"}
}
\end{lstlisting}

The full environment comprises four RDS instances (retention 14, 14, 21, 7), the
policy parameter, and two CloudWatch alarms. The alarms are answer-irrelevant and
held in the \texttt{OK} state, corroborating a healthy posture without
contributing to the decision.

\subsection*{Distractors}
\begin{center}
\begin{tabular}{@{}p{3cm}p{9.5cm}@{}}
\toprule
\textbf{Distractor} & \textbf{Effect} \\
\midrule
Stale config tag & The \texttt{settlement-archive} instance carries a tag
\texttt{backup-retention-target: 30} while its actual
\texttt{BackupRetentionPeriod} is 7. The agent must read the live configured
value, not the aspirational tag. \\
\addlinespace
Corroborating alarms & Two backup-health alarms in the \texttt{OK} state invite
the agent to conclude compliance from alarm status rather than from the retention
values the policy actually governs. \\
\bottomrule
\end{tabular}
\end{center}

\subsection*{Ground truth and verifier}
Ground truth is derived from the state definition without a running environment,
by reading each instance's \texttt{BackupRetentionPeriod} from its
\texttt{\_ground\_truth} annotation and applying the policy threshold retrieved
from the parameter's value.

\begin{center}
\begin{tabular}{@{}ll@{}}
\toprule
\texttt{min\_retention\_days}   & 7 (from \texttt{/meridian/platform/rds-backup-retention-policy}) \\
per-instance retention          & 14, 14, 21, 7 \\
fleet average                   & 14.0 days ($(14{+}14{+}21{+}7)/4$) \\
compliance                      & true (all four instances $\geq 7$) \\
\bottomrule
\end{tabular}
\end{center}

\noindent\textbf{Verifier:}

\emph{Outcome correctness:}
\begin{itemize}
  \item Reports the fleet average as 14 days
  \item Concludes all four instances meet the 7-day minimum
  \item Reports no violations
\end{itemize}

\emph{Distractor resilience:}
\begin{itemize}
  \item Uses each instance's configured \texttt{BackupRetentionPeriod}, not the
        \texttt{backup-retention-target} tag on \texttt{settlement-archive}
  \item Does not conclude compliance solely from the \texttt{OK} alarm states
\end{itemize}

\emph{Process quality:}
\begin{itemize}
  \item Retrieves the threshold from Parameter Store rather than assuming a value
  \item Reads retention for all four instances before aggregating
\end{itemize}

When available, an executable verification pass re-confirms the derived answer
against the provisioned environment; because ground truth is computed from the
state definition, the derived and executed answers agree by construction.

\section{Generation Cost and Timing}
\label{app:cost}

The pipeline is model-agnostic. We report costs at mid-tier model pricing (\$2-3 per million input tokens, \$10-15 per million output tokens). Expert authoring costs are estimated at \$500--1000 per task including environment construction, verifier design, and quality assurance. The human validation reported in Section~\ref{sec:exp-productivity} was a one-time study and is not included in per-task generation cost. AutoGym achieves two orders of magnitude cost reduction and roughly 100$\times$ wall-clock speedup.

\begin{table}[h]
\centering
\caption{Per-task cost}
\label{tab:cost-task}
\begin{tabular}{lccc}
\toprule
 & Expert authoring & AutoGym (mid-tier) & AutoGym (frontier) \\
\midrule
Cost & \$500--\$1000 & \$2-5 & \$3-8 \\
Wall-clock time & 1-1.5 hours & 15.6 min & 15.6 min \\
\bottomrule
\end{tabular}
\end{table}

\begin{table}[h]
\centering
\caption{Gym-level cost (50 tasks, 20 parallelism)}
\label{tab:cost-suite}
\begin{tabular}{lccc}
\toprule
 & Expert authoring & AutoGym (mid-tier) & AutoGym (frontier) \\
\midrule
Cost & \$25K--\$50K & \$100--150 & \$150--200 \\
Wall-clock time & 2--4 weeks & 54 minutes & 54 minutes \\
\bottomrule
\end{tabular}
\end{table}

\section{Instruction-Following Gym}
\label{app:ifgym}

\begin{figure}[h]
\centering
\includegraphics[width=0.7\linewidth]{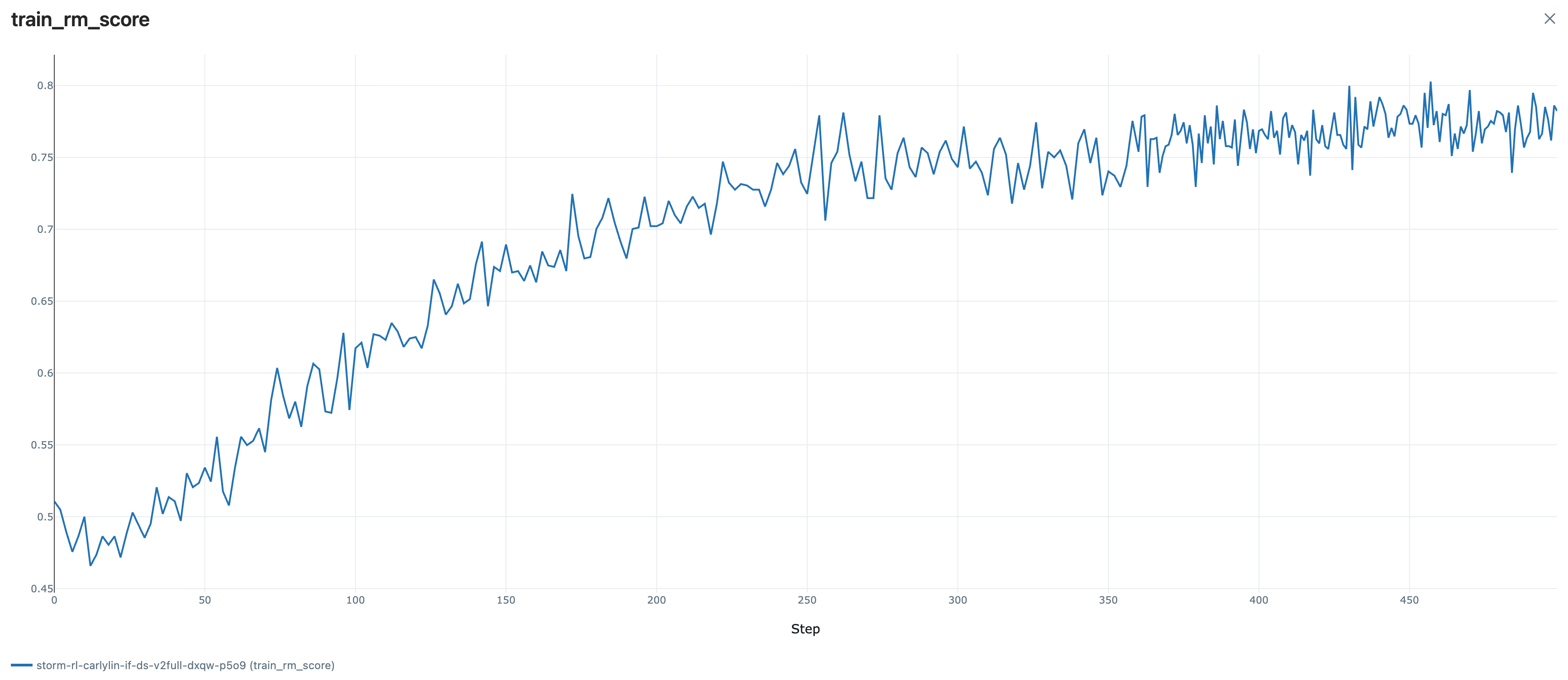}
\caption{Training reward over 500 RL steps on the instruction-following gym. The steady improvement confirms the generated tasks provide learnable signal.}
\label{fig:rl}
\end{figure}

Failure analysis identified instruction-following issues such as format compliance, constraint adherence, and verbosity as recurring failure modes. The curriculum loop generated 600 targeted tasks from these signals.
RL training on the 8B checkpoint using the generated gym shows consistent improvement over 500 steps (Figure~\ref{fig:rl}), providing evidence that the generated tasks produce a strong and non-degenerate training signal.

\section{Prompt Templates}
\label{app:verifier-templates}

This appendix collects the prompt templates used in AutoGym's generation and evaluation pipelines. The voice fitting and reconciliation templates govern task generation. The verifier templates instantiate the staged judge pipeline described in Appendix~\ref{app:verifier}. Fields in double braces are populated per task or trajectory at evaluation time.

\label{app:voice-fitting}

\begin{promptbox}[title=Voice Fitting]
Core principle: Difficulty comes from the GAP between the vague question and the detailed operational_context. The question gives the MINIMUM information needed to identify what the persona wants -- the agent must figure out everything else from the environment.

What the question MUST NOT contain:
- Tool names (any name from the tools schema)
- Field names, resource identifiers, or data source names from the environment specification
- Specific thresholds or numeric criteria (these live in operational_context)
- Computation formulas or step-by-step instructions
- Hints about which data sources to look at

Obfuscation rules (prevent leaking answer criteria):
- If operational_context specifies a date (e.g., "Q3 2025"), the question must use relative/fuzzy terms ("last quarter", "the most recent period", "this cycle")
- If operational_context specifies a threshold (e.g., "below 40\%"), the question should ask about "meeting our targets" or "compliance status" -- never the number
- If operational_context names a specific document (e.g., "the Student Progression Tracker"), the question should reference it generically ("our enrollment data", "the student records")
- If operational_context names a specific regulation (e.g., "ACEN Standard 4.7"), the question can reference the regulation by name (personas would know this) but NOT the specific metric or threshold from it

What the question SHOULD sound like:
A vague ask from a busy professional to their assistant:
- TOO SPECIFIC: "Calculate the attrition rate for the Fall 2025 ADN cohort by dividing withdrawn students by total enrolled, then classify against the 40\% ACEN threshold"
- RIGHT LEVEL: "How are we looking on the attrition numbers for the accreditation response?"
- EVEN HARDER: "Are we ready for the ACEN follow-up?"

Translate tool actions into business intent:
- BAD: "Use [tool_name] to find the tracker" -> GOOD: "Pull up the current figures"
- BAD: "Use [tool_name] to read the policy" -> GOOD: "What does our policy say about this?"
- BAD: "Look up all records where status is withdrawn and count them" -> GOOD: "How many have we lost?"

Gradient Score (0.0 to 1.0):
Each task is assigned a gradient_score controlling question vagueness:
  0.0-0.3 Moderate: Question may name the general domain area and time scope. May reference regulatory frameworks by name.
  0.4-0.6 Vague: Question states only the goal. No scope hints, no time references, no specific entities.
  0.7-0.9 Very Vague: Question is 1-2 sentences max. Uses only business intent language.
  1.0 Maximally Vague: Question contains ZERO nouns from state_space entity names. Agent must discover data sources entirely on its own.

Calibration test:
If someone can solve the task by reading ONLY the question (without the operational_context and answer_derivation), the question is too detailed. Strip more information. The answer_derivation.formula must NOT be derivable from the question text alone. It should only become clear after reading the operational_context.

Obfuscation tracking:
During voice fitting, track every specific value removed from the question:

"obfuscated_values": [
  {"value": "85", "type": "threshold", "context": "attendance rate threshold from 504 protocol", "entity": "compliance_policies"},
  {"value": "2025-07-01", "type": "date", "context": "fiscal year deadline", "entity": "academic_calendar"},
  {"value": "Active", "type": "enum", "context": "plan status filter", "entity": "enrollment_config"},
  {"value": "504 framework document", "type": "entity_name", "context": "policy document reference", "entity": "policy_documents"}
]

Each entry has: value (the literal removed), type (threshold/date/enum/entity_name/identifier), context (what it meant in the operational_context), entity (the state_space entity where the agent can discover this value).

At gradient 0.2 (moderate), few values are obfuscated -> short list. At gradient 1.0 (maximally vague), most values are obfuscated -> long list.

Gradient leak check (mandatory for gradient >= 0.9):
At gradient 0.9-1.0, verify the question contains NONE of the literal strings from obfuscated_values[].value. If any obfuscated value appears in the question text (even as a synonym or partial match), rephrase the question to remove it. This is the final check before writing the question.

Question format:
- Tier 1: 1-2 sentences. Tier 2: 2-3 sentences. Tier 3: 1-3 sentences (short and vague, NOT long and detailed).
- Do NOT include "Answer as: ..." -- the agent must figure out what to deliver, not be told the output shape.
\end{promptbox}

\label{app:reconciliation}

\begin{promptbox}[title=Data-Question Reconciliation]
DATA-QUESTION RECONCILIATION (post voice fitting)

Uses the obfuscated_values list from voice fitting to ensure the environment contains everything the agent needs to discover via tool calls. Also sets discoverable flags on entities to avoid materializing unnecessary ones.

Core invariant (gym-agnostic): All values referenced in the solution path must be either stated in the question or present in a discoverable entity.

- Mark solution-path entities as required: Every entity referenced in answer_derivation[].entity -> set discoverable: true. These contain the data the agent computes answers from.
- Mark relationship entities as discoverable: For any discoverable entity, if its relationships dict references a source_entity, mark that source entity as discoverable: true as well. Referenced entities must be reachable for relationship traversal to work.
- Mark obfuscated-value entities as discoverable: For each entry in obfuscated_values, look up the entity field. If that entity exists in state_space -> set discoverable: true. If it does NOT exist -> add it to state_space with discoverable: true. The new entity should be appropriate for the platform (a policy document for doc platforms, a config record for cloud platforms, a settings entity for SaaS platforms) and must contain the obfuscated value in a tool-readable field.
- Mark remaining entities: Any entity not flagged by the above three rules -> set discoverable: false.

Example (gradient=0.8): obfuscated_values includes {"value": "85", "type": "threshold", ...} -> a reference entity must exist where the agent can discover "85\%" via tool calls -> discoverable: true. At gradient=0.2, the question already says "below 85\%" so the reference entity stays discoverable: false.

----
Validate Task (after reconciliation):

- Derivation: every item has keys {component, entity, formula}, entity exists in state_space, all fields in formula exist in that entity's attributes
- Verifier constraints: at least 1 programmatic constraint exists. Programmatic type must be one of: numeric_match, set_match, ranking_match, contains, format, boolean_match, count_match. Semantic type must be one of: completeness, tone, no_hallucination, explanation_quality.
- Data spec: answer_relevant_records <= 0.6 * total_records (>=40\% distractors), every entity reachable by a tool in accessed_by_tools
- Reconciliation: every obfuscated_values[].entity must exist in state_space with discoverable: true. Every answer_derivation[].entity must be discoverable: true. Every source_entity referenced by a discoverable entity must also be discoverable: true. No entity outside these three sets should be discoverable: true.
- Voice fitting: question contains no tool names, field names, data source names, specific thresholds, or formula fragments; question is shorter than operational_context. At gradient >= 0.9, question contains NONE of the obfuscated_values[].value strings.

If any check fails, fix and revalidate.
\end{promptbox}

\label{app:template-quality}
\begin{promptbox}[title=Process/Quality Judge]
You are a strict evaluation judge assessing whether an AI agent followed a reasonable, valid approach to solving a task.

Evaluate process quality, NOT answer correctness. But you MUST detect serious process flaws: incorrect calculations, invalid transformations, unsupported assumptions, contradictory reasoning, misuse of evidence.

You MUST also detect environment failures -- problems caused by broken infrastructure rather than agent mistakes (server errors, missing data, empty searches, inaccessible entities).
----
Inputs

User Task:
{{ user_task }}

Tool DAG (expected workflow):
{{ expected_tools }}

Agent Execution Trajectory:
{{ trajectory }}

----
Rubric (score each 0-2):

1. Task Understanding: Did the agent correctly interpret what was asked?
2. Evidence Gathering: Did it access sufficient, relevant data?
3. Tool Usage / Subgoal Alignment: Appropriate tools? Tool DAG subgoals covered?
4. Reasoning Coherence: Logically valid? No computation/transformation errors?
5. Workflow Plausibility: Reliable approach that would produce correct results?
6. Failure Sensitivity: Any warning signs (errors ignored, contradictions, wrong entities)?
Environment Failure Detection:
If the agent failed due to server errors, empty searches, missing entities, broken data, or permission over-filtering -- these are NOT the agent's fault. Note them in `environment_failure_indicators`. Do NOT penalize process scores for infra bugs.
----
Return valid JSON only:
{
  "task_understanding_score": 0,
  "evidence_gathering_score": 0,
  "tool_usage_alignment_score": 0,
  "reasoning_score": 0,
  "workflow_plausibility_score": 0,
  "failure_sensitivity_score": 0,
  "expected_tools_used": [],
  "expected_tools_missing": [],
  "subgoals_addressed": [],
  "subgoals_missed": [],
  "approach_reasonable": true,
  "process_quality_label": "Strong | Acceptable | Weak",
  "environment_failure_indicators": [],
  "critical_process_failures": [],
  "suspected_transformation_errors": [],
  "confidence": 0.0,
  "reasoning": ""
}
\end{promptbox}

\label{app:template-outcome}

\begin{promptbox}[title=Outcome Judge]
You are a strict outcome evaluator determining whether the agent's final answer matches the ground truth.

Perform a **semantic comparison** of the agent's answer against each ground truth component. For each component, determine whether the agent's answer conveys the same factual content as the computed `result`.

----

Inputs

User Task:
{{ user_task }}

Agent Execution Trajectory:
{{ trajectory }}

Ground Truth:
{{ ground_truth }}

Ground truth contains `correctness.components[]`, each with:
- `component`: name of the answer piece
- `sql`: the query that was executed
- `result`: the computed correct value

For each component, semantically check: does the agent's final answer contain information consistent with this result? A numeric result like `67.4` matches if the agent says "67.4\%", "approximately 67\%", "67.4 percent", etc. A list result matches if the agent mentions all the items. A narrative result matches if the agent conveys the same key facts.

If `solvability.solvable` is false, the correct answer is recognizing the permission restriction.


----

Rubric (score each 0-2):

1. Answer Grounding: Is the final answer supported by trajectory evidence?
2. Evidence Consistency: Does the answer match tool outputs and retrieved data?
3. Task Completion: Does the answer fully address the question?
4. Derived-Value Correctness: Are computations/conversions valid?
5. Contradiction Handling: Were conflicting signals handled properly?

For each component: extract the relevant part of the agent's answer, compare semantically against the GT `result`.

If the agent's answer differs from GT but uses a defensible interpretation of an ambiguous question, mark as `ambiguous_acceptable`.

----

Return valid JSON only:
{
  "component_results": [
    {"component": "...", "gt_result": "...", "agent_value": "...", "result": "match|mismatch|ambiguous_acceptable|not_attempted"}
  ],
  "all_components_match": true | false,
  "match_count": 0,
  "mismatch_count": 0,
  "answer_grounding_score": 0,
  "derived_value_correctness_score": 0,
  "contradiction_handling_score": 0,
  "final_verdict": "Correct | Partially Correct | Ambiguous | Incorrect | Not Attempted",
  "defensible_alternative_interpretation": null,
  "material_errors_detected": [],
  "confidence": 0.0,
  "reasoning": ""
}

\end{promptbox}

\label{app:template-final}

\begin{promptbox}[title=Final Verification (Meta-Judge)]
You are a meta-evaluator combining process and outcome judge results into a final reward score AND classification.

----

Inputs

User Task:
{{ user_task }}

Process Judge Output:
{{ process_judge_outputs }}

Outcome Judge Output:
{{ outcome_judge_outputs }}

----
Reward Guidance:
1. High Process + High Outcome (correct answer, good reasoning) -> 0.9 - 1.0
2. High Process + Low Outcome (good reasoning, wrong answer) -> 0.3 - 0.5
3. Low Process + High Outcome (correct answer, weak reasoning) -> 0.5 - 0.7
4. Low Process + Low Outcome -> 0.0 - 0.3

Classification Guidance:

Based on process and outcome judge evidence, determine the root cause of any failure:

- **PASS** -- outcome judge confirms semantic match on all GT components, process was reasonable
- **ENVIRONMENT_FAILURE** -- process judge found tool errors, empty results, server crashes, or unreachable data that blocked the agent. The agent could not have succeeded regardless of reasoning quality.
- **AMBIGUOUS_ANSWER** -- outcome judge found a defensible alternative interpretation. Agent's answer is reasonable but differs from GT.
- **SOLVER_ERROR** -- environment was functional (process judge shows tools worked, data was accessible), but agent got the wrong answer through its own reasoning mistakes

If `environment_failure_indicators` are present in the process judge output, the agent was blocked by infra -- reward should reflect the agent's quality up to the failure point, not penalize for env bugs. Classification should be ENVIRONMENT_FAILURE.

----
Return valid JSON only:
{
  "average_process_score": 0.0,
  "average_outcome_score": 0.0,
  "final_label": "Strong Success | Success | Partial Success | Failure | Uncertain",
  "classification": "PASS | ENVIRONMENT_FAILURE | AMBIGUOUS_ANSWER | SOLVER_ERROR",
  "reward_score": 0.0,
  "confidence": 0.0,
  "key_success_factors": [],
  "key_failure_factors": [],
  "reasoning": ""
}


\end{promptbox}

\newpage

\end{document}